\documentclass{article}

\usepackage{PRIMEarxiv}

\usepackage[utf8]{inputenc} 
\usepackage[T1]{fontenc}    
\usepackage{hyperref}       
\usepackage{url}            
\usepackage{booktabs}       
\usepackage{amsfonts}       
\usepackage{nicefrac}       
\usepackage{microtype}      
\usepackage{lipsum}
\usepackage{fancyhdr}       
\usepackage{graphicx}       
\usepackage{tabularx}
\newcolumntype{C}{>{\centering\arraybackslash}X}
\usepackage{multirow}
\usepackage{graphics}
\usepackage{float}
\usepackage{verbatim} 
\usepackage{tikz}
\usepackage{amsmath,amssymb} 
\usepackage{commath}

\title{Pushing the Envelope for Depth-Based Semi-Supervised 3D Hand Pose Estimation with Consistency Training}

\author{
  Mohammad Rezaei\\
  Department of Computer Science and Engineering \\
  University of Texas at Arlington \\
  Arlington\\
  \texttt{mohammad.rezaei@mavs.uta.edu} \\
   \And
  Farnaz Farahanipad \\
  Department of Computer Science and Engineering \\
  University of Texas at Arlington \\
  Arlington\\
  \texttt{farnaz.farahanipad@mavs.uta.edu} \\
   \And
  Alex Dillhoff \\
  Department of Computer Science and Engineering \\
  University of Texas at Arlington \\
  Arlington\\
  \texttt{alex.dillhoff@uta.edu} \\
   \And
  Vassilis Athitsos \\
  Department of Computer Science and Engineering \\
  University of Texas at Arlington \\
  Arlington\\
  \texttt{athitsos@uta.edu} \\
}

\begin{document}
\maketitle

\begin{abstract}
Despite the significant progress that depth-based 3D hand pose estimation methods have made in  recent years, they still require a large amount of labeled training data to achieve high accuracy. However, collecting such data is both costly and time-consuming. To tackle this issue, we propose a semi-supervised method to significantly reduce the dependence on labeled training data. The proposed method consists of two identical networks trained jointly: a teacher network and a student network. The teacher network is trained using both the available labeled and unlabeled samples. It leverages the unlabeled samples via a loss formulation that encourages estimation equivariance under a set of affine transformations. The student network is trained using the unlabeled samples with their pseudo-labels provided by the teacher network. For inference at test time, only the student network is used. Extensive experiments demonstrate that the proposed method outperforms the state-of-the-art semi-supervised methods by large margins.
\end{abstract}

\keywords{Deep Learning, Computer Vision, 3D hand pose estimation}

\section{Introduction}
The hands are the primary means by which humans interact with the outside world. As such, accurate hand pose estimation is a necessary requirement for many vision-based systems and enables many applications in areas such as augmented reality (AR), virtual reality (VR) and gesture recognition.
\par
Recently, the availability of more accurate and affordable commodity depth cameras coupled with the success of Deep Neural Networks (DNN) has led to significant progress in depth-based 3D hand pose estimation and segmentation \cite{yuan2018depth,malik2020handvoxnet,chen2019so,du2019crossinfonet,xiong2019a2j,ge2018point,wan2018dense,baek2018augmented,tompson2014real,bojja2019handseg}. Despite these advancements, one major challenge that remains is that DNN-based methods require large amounts of annotated training data to realize their full potential. A straightforward approach to mitigate this requirement is to use synthesized training data with accurate annotations \cite{poier2019murauer}, which can be generated with minimal human effort. However, models trained on synthesized data generalize poorly to the real-world data due to the significant domain gap between synthetic and real-world data. A popular alternative is semi-supervised learning (SSL) \cite{chapelle2009semi}, where the goal is to leverage unlabeled data along with the labeled data, hence reducing the amount of labeled data required for training. Most of the recent advancements of SSL methods have been focused on image classification \cite{demiriz1999semi,gammerman2013learning,joachims1999transductive,grandvalet2005semi,blum1998combining,nigam2000analyzing,belkin2004regularization,blum2001learning,he2018amc,wang2007label,li2021comatch}.
\par
Semi-supervised learning has recently attracted attention in the area of 3D hand pose estimation. Chen et al.~\cite{chen2019so} leverage unlabeled data by minimizing the Chamfer loss between the input point cloud and its reconstructed version by a decoder. Wan et al.~\cite{wan2017crossing} jointly train two deep generative models with a shared latent space to model the statistical relationships of depth images and their corresponding hand poses. Their architectural design facilitates learning from unlabeled data. Poier et al.~\cite{poier2019murauer} exploit synthetic data to reduce reliance on annotated real-world data by learning to map from the features of real data to that of synthetic data. Baek et al.~\cite{baek2018augmented} synthesize data in the skeleton space and then its corresponding depth map to augment the training data. These methods enable semi-supervised learning through accommodations in their network architecture. Orthogonal to this line of work, we propose a model-agnostic semi-supervised framework for 3D hand pose estimation that takes advantage of the most recent advancements of SSL methods in image classification.
\par
The proposed framework consists of two identical networks that are trained jointly: 1) student network and 2) teacher network. Any off-the-shelf network architecture can be used as long as it provides a means for prediction uncertainty estimation. For training the teacher, we adopt an approach based on consistency training \cite{xie2019unsupervised}. Driven by the intuition that a good model should be robust to any small change in an input example, approaches based on the consistency training enforce the model predictions to be invariant to small noise applied to input examples. Inspired by this approach, we train the teacher network using both the labeled and unlabeled parts of the training data, with a combination of the typical supervised loss and an unsupervised loss formulated in such a way to enforce model consistency defined as the model output equivariance under a set of affine transformations. 
\par
Note that the proposed method uses different training strategies from \cite{xie2019unsupervised} due to the fundamentally different nature of the 3D hand pose estimation, which is a structured regression task as opposed to an image classification task. This difference poses unique challenges for a hand pose estimation method based on consistency training, necessitating not only architectural changes but also different training strategies. We present several novel components to effectively address these challenges. The student network is trained using the pseudo-labels generated by the teacher network. More specifically, to stabilize the training, exponential moving average \cite{cai2021exponential} of the teacher network's parameters are used for generating the pseudo-labels. After the training is finished, the student network is fine-tuned on the labeled part of the training data since it has not seen any of them during training. Note that the proposed method comes at no additional cost at test time, as only the student network is used for inference and the teacher network is discarded after training.
\par
 It should be stressed that the proposed training of a separate student network is different from knowledge distillation~\cite{hinton2015distilling}. The goal of knowledge distillation is to transfer the knowledge of a complicated model to a simpler model by training the simpler model with the softmax outputs of the complicated model. Moreover, knowledge distillation is performed after training. However, the proposed method employs two identical networks that are trained simultaneously. Furthermore, the student is trained using the exponential moving average (EMAN) of the teacher network's parameters. 
\par
We conduct an extensive evaluation of the proposed method on three publicly available datasets, namely ICVL~\cite{tang2014latent}, MSRA~\cite{sun2015cascaded} and NYU~\cite{tompson2014real}, which are challenging benchmarks commonly used for evaluation of 3D hand pose estimation methods. The results demonstrate that the proposed method significantly outperforms the current state-of-the-art semi-supervised hand pose estimation methods. We also analyse the performance of the proposed method in cases of severe scarcity of ground-truth annotations and show its effectiveness under such scenarios. Most remarkably, using only 25\% of the annotations, the proposed method preforms on par with the state-of-the-art fully supervised methods (methods that use 100\% of the ground-truth annotations).
\par
In summary, our contributions are as follows:
\begin{itemize}
  \item We propose a novel semi-supervised hand pose estimation method to effectively leverage the unlabeled data. To the best of our knowledge, this is the first method to incorporate consistency training for semi-supervised training on depth images of hands. 
  \item We propose several novel strategies to enable consistency training for 3D hand pose estimation on depth images.
  \item The proposed method is the first depth-based hand pose estimation method to incorporate advances from recent SSL methods such as~\cite{sohn2020fixmatch,xie2019unsupervised,cai2021exponential}, which target general-purpose image classification. A key contribution is proposing concrete ways to apply those ideas to depth-based hand pose estimation, and showing that they lead to improved performance.
  \item We empirically show that the proposed method outperforms the current state-of-the-art semi-supervised 3D hand pose estimation methods.
\end{itemize}
Code and models will be made publicly available upon acceptance.

\section{Related Work}
\subsection{Hand Pose Estimation}
Hand pose estimation has been a long-standing problem in the Computer Vision community. While early methods relied on non-data-driven approaches such as hand crafted features, optimization methods, and distance metrics \cite{athitsos2003estimating,sharp2015accurate,oikonomidis2011efficient}, in recent years there has been a shift to methods based on deep neural networks (DNNs). While there have been many DNN-based methods proposed to perform 3D hand pose estimation from different data modalities, we keep our focus here on methods designed for depth images, as they are the most related to ours.
\par
Oberweger et al.~\cite{oberweger2015hands} proposed a method to estimate the hand pose represented by PCA coefficients of a statistical hand model. Wang et al.~\cite{wang2018region} use the ensemble principle by partitioning the last convolutional outputs of a CNN into several regions and using separate regressors to estimate the hand joints.  Another line of work is to take advantage of the recent advancements in 3D Deep learning. To this end, \cite{ge20173d,moon2018v2v} converted 2.5D depth images into 3D voxels and employed 3D CNNs to estimate the 3D hand pose. Several methods  \cite{ge2018hand,ge2018point,li2019point,chen2019so,cheng2021handfoldingnet,huang2020hand} have been recently proposed to utilize point cloud processing networks by converting depth images into point clouds as the input. Comprehensive reviews of depth-based hand pose estimation can be found in \cite{supancic2015depth,yuan2018depth}. 
\par
The above-mentioned methods are all fully-supervised. Our work is most related to the recent line of semi-supervised methods for 3D hand pose estimation \cite{chen2019so,wan2017crossing,poier2019murauer,baek2018augmented}. SemiHand~\cite{yang2021semihand} is closest to our method. It uses consistency training, which is also the case for the proposed method. However, the consistency training approach in the proposed method is significantly different than that of ~\cite{yang2021semihand}. Besides using a different input data modality (2.5 depth images as opposed to RGB images), the proposed framework is based on a student-teacher paradigm and uses consistency training only for training the teacher network, whereas SemiHand~\cite{yang2021semihand} uses a single network. The proposed framework also uses a fundamentally different mechanism for label-refinement and the network uncertainty estimation. Finally, unlike SemiHand~\cite{yang2021semihand}, the proposed method only uses view consistency.

\subsection{Semi-supervised Learning in Image Classification}
The key challenge to training of modern DNNs is the requirement for large amounts of labeled data. Semi-supervised learning (SSL) mitigates this requirement by providing a means of leveraging unlabeled data. Classic examples of SSL methods include transductive models \cite{demiriz1999semi,gammerman2013learning,joachims1999transductive}, entropy minimization \cite{grandvalet2005semi}, co-training \cite{blum1998combining,nigam2000analyzing} and graph-based models \cite{belkin2004regularization,blum2001learning,he2018amc,wang2007label,li2021comatch}.
\par
Our work is closely related to the recent line of SSL methods based on pseudo-labeling \cite{arazo2020pseudo,han2019deep,cascante2020curriculum}, where they produce artificial label for unlabeled data samples and train the model to predict the artificial label when fed unlabeled samples as input, and consistency training \cite{sohn2020fixmatch,xie2019unsupervised,gong2021alphamatch} wherein they enforce the model predictions to be consistent across a sample and its perturbed version. However, the proposed method is fundamentally different from the methods discussed above. They are all focused on image classification, where the goal is encourage representation invariance across different views of the same image. However, the proposed method performs hand pose estimation, which is a structured regression task. It critically depends on spatial information and its goal is to enforce representation equivariance across different views. These differences pose unique challenges for a hand pose estimation method based on consistency training. In this paper, we propose several novel strategies to address these challenges and take advantage of the state-of-the-art SSL methods in image classification.

\section{Proposed Method}
\subsection{Problem Formulation and Notation}
The task of 3D hand pose estimation is defined as follows: given an input depth image $x\in{\mathbb{R}^{H\times{W}}}$, the task is to estimate the 3D location of a set of pre-defined hand joints $\mathcal{J}\in{\mathbb{R}^{N_J\times{3}}}$ in the camera coordinate system by learning a mapping $f$ in the form of a neural network parameterized by $\theta$, such that $\mathcal{J} = f(X ;\theta)$. $H$ and $W$ denote the height and width of the depth map respectively. For the sake of simplicity, we refer to the input data dimensionality as $d = H\times{W}$. We use $N_J$ to refer to the number of estimated joints. $\mathcal{J}_i = (U,V,Z)$ represents the location of the the $i^{th}$ joint .The function $f$ is learned using the training set consisting of labeled examples $(x_l, \mathcal{J}^l) \sim P_L$  and unlabeled examples $x_u \sim P_U$. $P_l$ and $P_U$ denote the probability distributions of labeled and unlabeled examples respectively. We define an augmentation function $\Phi: \mathbb{R}^{d} \rightarrow{\mathbb{R}^{d}}$ such that it maintains the equivariance property. This mathematically means that if $x^{\prime} = \Phi(x)$, then we have $\mathcal{J}^{\prime} = \Phi(\mathcal{J})$. We define $\mathcal{M}$ as a uniform probability distribution over all such augmentation functions. In our experiments, we use a subset of affine transformations including translation, scaling and rotation as augmentation functions. 
\par
\begin{figure}[t]
\centering
\includegraphics[height=5.2cm]{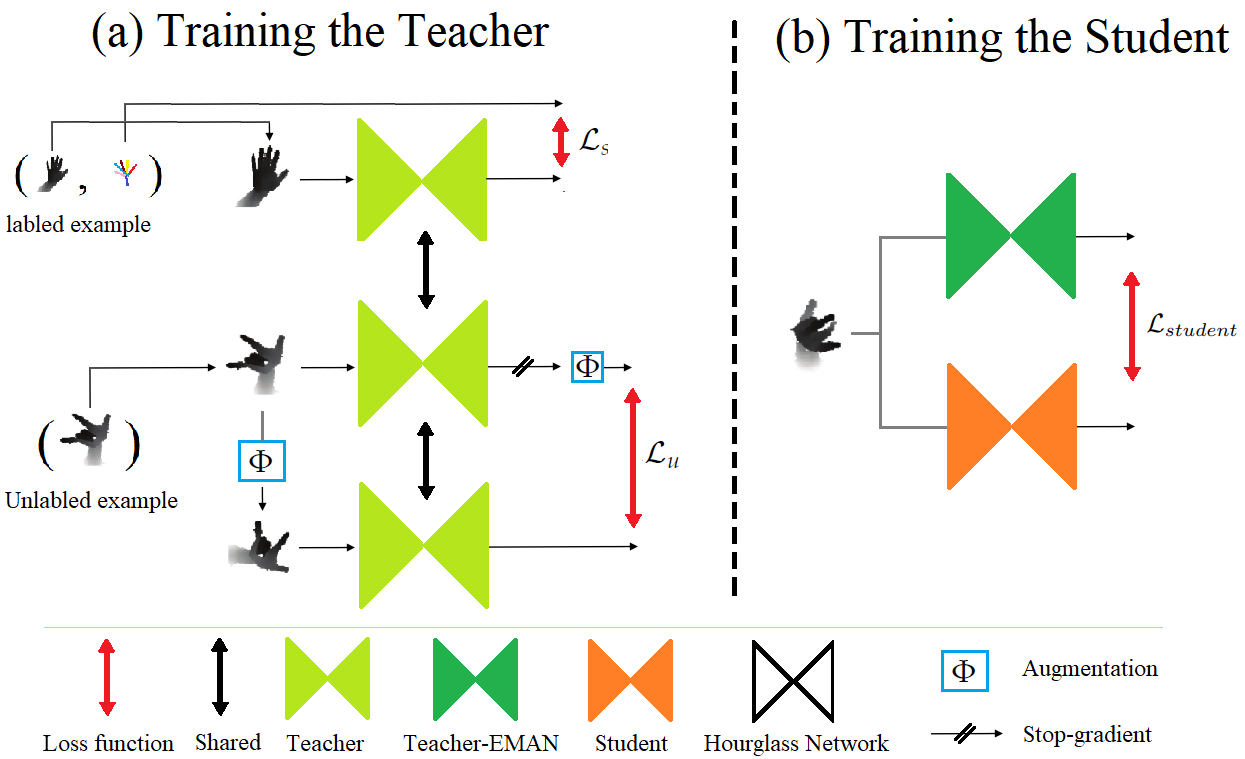}
\caption{Left: overview of training the teacher network. It depicts a training batch consisting of one labeled example and one unlabeled example. The teacher network is trained using a combination of the typical supervised loss on the labeled examples and consistency loss on the unlabeled examples. Right: the student network is trained using the pseudo-labels provided by the EMAN of the teacher network's parameters}
\label{fig:overview}
\end{figure}

The overview of the proposed method is illustrated in Fig. \ref{fig:overview}. It employs two identical networks called the teacher network and the student network, whose parameters are denoted by $\theta_T$ and $\theta_S$ respectively. The teacher network task is to provide supervisory signal for the student network by generating pseudo-labels. It is trained using a combination of the typical supervised loss and consistency training loss. As the teacher network improves, so do the pseudo-labels it generates for the student network. As a result, the student network keeps improving as the training of the teacher network progresses. After training is finished, the student network will be fine-tuned using the available labeled samples because it has not seen any of them during training. We empirically found that this leads to some modest performance improvements.

\subsection{Network Architecture}
Both the teacher and student network follow the same architecture that is similar to \cite{iqbal2018hand}. It consists of an encoder network and two separate branches. The encoder is a CNN whose task is to extract hand features from the input depth image. Its output feature volume serves as the input to two branches. The first branch estimates a heatmap $H_j^{2D}$ for each joint.  The 2D heatmap $H_j^{2D}$ represents the occurrence likelihood of the $j^{th}$ joint at each pixel location. The second branch estimates a depth map $H_j^z$ for each joint. $H_j^z$ represents depth prediction for the corresponding pixels for the $j^{th}$ joint. The 3D locations are then computed following \cite{sun2018integral,iqbal2018hand}:

\begin{equation}
  (U^j,V^j) = \sum_{u_i}\sum_{v_i}(u_i,v_i){\hat{H}}_j^{2D}(u_i,v_i)
\end{equation}
\begin{equation}
  Z^j = \sum_{u_i}\sum_{v_i}H_j^z(u_i,v_i){\hat{H}}_j^{2D}(u_i,v_i)
\end{equation}

In the above, ${\hat{H}}_j^{2D}(u_i,v_i)$ is the $H_j^{2D}$ normalized through spatial Softmax operation as follows:
\begin{equation}
  {\hat{H}}_j^{2D}(x,y) = \frac{exp({\alpha}_j H_j^{2D}(x,y))}{\sum\limits_{u_i,v_i\in{\Omega}}exp({\alpha}_j H_j^{2D}(u_i,v_i))}
\end{equation}
Here, $\Omega$ represents the set of all pixel locations in the input map $H_j^{2D}$. ${\alpha}_j$ denotes the temperature parameter that controls the spread of the output heatmaps ${\hat{H}}_j^{2D}$. Unlike \cite{iqbal2018hand} that trains these parameters along with the rest of the network parameters, we set ${\alpha}_j = 1$ for $j \in{1,2,...,N_J}$ and keep them fixed during the training phase. In our experiments, we use an Hourglass network \cite{newell2016stacked} as the encoder.


\subsection{Teacher Network Training}
The teacher network is trained using both labeled and unlabeled examples by solving the following optimization problem, conceptually similar to \cite{xie2019unsupervised,yang2021semihand}:  

\begin{equation}
\label{eqn:teacher}
\begin{split}
  \min_{\theta_T} \mathcal{L}_{s}(\theta_T) + \lambda \mathcal{L}_{u}(\theta_T) =&  \mathbb{E}_{(x_l,\mathcal{J}^l)\sim P_L(x)}[D(f(x_l ;\theta_T),\mathcal{J}^l)] + \\
 & \lambda \mathbb{E}_{x_u\sim P_U(x)}\mathbb{E}_{\Phi\sim \mathcal{M}}[D(\Phi(f(x_u ;\bar{\theta_T})),f(\Phi(x_u) ;\theta_T))] 
\end{split}
\end{equation}

Here, $D$ denotes mean element-wise L1 distance. $\bar{\theta_T}$ denotes a fixed copy of the current parameters $\theta_T$, indicating the the gradient is not back-propagated through $\bar{\theta_T}$, as done in \cite{xie2019unsupervised,miyato2018virtual}. $\lambda$ is a weighting factor to balance the terms $\mathcal{L}_{s}$ and $\mathcal{L}_{u}$. $\mathcal{L}_{s}$ is the typical supervised loss computed on the labeled examples, which is aimed at minimizing the difference between the model predictions and the corresponding ground-truth labels. $\mathcal{L}_{u}$ is the unsupervised consistency regularization loss computed on unlabeled examples to enforce consistency of the model predictions across different views of the same depth images. In contrast to image classification where the consistency is defined as the model prediction invariance across different views \cite{xie2019unsupervised}, we define consistency as equivariance under a set of affine transformations, similar to SemiHand~\cite{yang2021semihand}. Explicitly enforcing such an estimation equivariance on the unlabeled examples proves to be a very effective means of leveraging them. $f(x_u ;\bar{\theta_T})$ can be interpreted as the pseudo-labels generated by the teacher network from the view $x_u$ to be used by its own in the second view $\Phi(x_u)$.


\par
\textbf{Sample Masking.} Models trained using self-generated pseudo-labels generally suffer from confirmation bias \cite{tarvainen2017mean}, where the model keeps amplifying its own errors. It has been demonstrated that masking out noisy pseudo-labels and maintaining only high-quality ones for training can considerably reduce the confirmation bias~\cite{arazo2020pseudo}. The standard approach towards this end has been to base the decision of whether to use a pseudo-label for training on a model prediction certainty(or uncertainty) measure compared against a threshold. The typical approach in image classification, namely taking the maximum of the model output probability, is not applicable to the proposed method due to the different nature of its task. SemiHand~\cite{yang2021semihand} defined the model confidence on a data sample as the sum of the distance between the model's prediction on an image and that of its randomly perturbed version, and the distance between the pseudo-label and its corrected pseudo-label. However, this approach requires additional model evaluations and performing forward kinematic chain, which adds computational overhead. The proposed method uses a simple yet effective method to measure the model uncertainty. The prediction uncertainty for the $j^{th}$ joint, denoted by $C_j$, is approximated using the Standard deviation (STD) of the corresponding estimated normalized heatmap ${\hat{H}}_j^{2D}$, computed as follows:
\begin{equation}
  C_j = \sqrt{ \sum_{u_i}\sum_{v_i}{\hat{H}}_j^{2D}(u_i,v_i) \norm{[U_j,V_j]^T-[u_i,v_i]^T}^2 }
\end{equation}
Here, $\norm{.}$ denotes the Frobenius norm function. We empirically found that when the model is certain about its prediction on a given joint, its corresponding heatmap has a low STD. On the other hand, when the model is not certain and there are many candidate pixels, the heatmap tends to be wider (and hence high STD).
\par
We define the mask $M \in{\mathbb{R}^J}$ as follows:

\begin{equation}
\label{eqn:mask}
  M_j=
 \begin{cases}
    m_a & \text{if }C_j < T_j\\
    m_r               & \text{otherwise}
\end{cases}
\end{equation}

Here, $T_j$ is the threshold used for masking the $j^{th}$ joint. Symbols $m_a$ and $m_r$ denote the weights given to the pseudo-labels that are respectively accepted and rejected. Image classification SSL methods such as  \cite{sohn2020fixmatch,xie2019unsupervised,zhang2021flexmatch} have traditionally taken a binary approach for masking ($m_a = 1$ and $m_r = 0$), which is a special case of Eq. \ref{eqn:mask}. While this binary approach has proven effective for image classification, we empirically found that including 
the pseudo-labels that are rejected in the training ($m_r \neq 0$) consistently leads to performance improvement in the proposed method. 
The L1 distance $D$ in $\mathcal{L}_{u}$ in the Eq. \ref{eqn:teacher} is replaced by the following weighted average:

\begin{equation}
  D(\mathcal{J},\mathcal{J}^{\prime}) = \frac{1}{3K}\sum\limits_{j=1}^{N_J}{M_j \sum\limits_{i=1}^{3}{|\mathcal{J}_{ji} - \mathcal{J}^{\prime}_{ji}|}}
\end{equation}

where $K = \sum_j{M_j}$. In the case where masking is not used, we set all $M_j = 1$.
\par

\textbf{Dynamic Thresholding.} A standard practice of SSL methods in image classification is to use a fixed threshold for masking \cite{sohn2020fixmatch,xie2019unsupervised}. Most recently, \cite{zhang2021flexmatch} employed a strategy for dynamic adjustment of thresholds for different classes based on class learning effects. However, none of the these strategies is practical for the proposed method. Naively using fixed thresholds for masking causes two issues in the proposed method. First, since the the uncertainty measure for different joints are of different scales (e.g. heatmaps corresponding to fingertips are usually very peaky but are relatively wide for the palm, leading to low and high STDs respectively), we would need $N_J$ different thresholds (one for each joint), making the hyper-parameter optimization complicated. Secondly, adopting uncertainty-based pseudo-labeling leads to a class imbalance in the pseudo-labels, and thereby, misguides the training \cite{nassar2021all}. 
\par
To tackle these issues, we employ a strategy to dynamically adjust the thresholds. Let ${\rho}_t$ be the fraction of pseudo-labels allowed for training at the training epoch $t$. Let $T^t_j$ be the threshold value used for masking for the $j^{th}$ joint at the training epoch $t$. After initialization in the first epoch, $T^{t+1}_j$  for the epoch $t+1$ is computed as follows:
\begin{equation}
\label{eqn:dynamicthreshold}
  T^{t+1}_j = T^t_j + \eta({\rho}_t - {\rho^j_t} )
\end{equation}
Here, ${\rho^j_t}$ is the fraction of pseudo-labels corresponding to the $j^{th}$ joint accepted for training in the epoch $t$ according to the corresponding threshold $T^t_j$. Symbol $\eta$ denotes the adjustment rate. Intuitively, when ${\rho}_t > {\rho^j_t}$, it means that the threshold $T^t_j$ should increase to let pass more pseudo-labels. On the other hand, when ${\rho}_t < {\rho^j_t}$, it means the threshold $T^t_j$ should decrease to accept fewer pseudo-labels for training. This type of addressing class-imbalance problem can be thought of as equivalent to Mean Sampling \cite{he2021rethinking}. It ensures that a roughly equal fraction ${\rho_t}$ of pseudo-labels for each joint is used for training in each epoch. We use a cosine schedule strategy \cite{loshchilov2016sgdr} to increase  ${\rho}_t$ from the initial value ${\rho}_{start}$ to reach  its final value ${\rho}_{end}$ over the course of training as follows:
\begin{equation}
\label{eqn:rhoo}
  {\rho}_{t} = {\rho}_{start} + 0.5 (1-\cos(\frac{T_{cur}}{T_{max}}\pi )) ({\rho}_{end} - {\rho}_{start})
\end{equation}
Here, $T_{cur}$ and $T_{max}$ denote the current epoch number and the total number of training epochs respectively. Intuitively, the proposed method allows only a small proportion of pseudo-labels to be used at early phases of the training since the teacher network is still not accurate in early training phases. As the training progresses and the teacher network performance improves, a larger proportion of pseudo-labels are allowed to be used for training. 
\par


\subsection{Student Network Training}
The proposed method trains the student network using the unlabeled samples and their corresponding pseudo-labels generated by the teacher network. However, using pseudo-labels generated directly by the teacher network can lead to a potential problem. The teacher itself is constantly updated in the training, which could cause performance degradation and training instability in the student network since it has to learn to approximate a highly non-stationary function. 
An alternative is to use the exponential moving average (EMA) of the teacher network's parameters to generate pseudo-labels \cite{tarvainen2017mean}. However, this approach could lead to a potential mismatch between the EMA parameters and the batch normalization (BN) statistics in the parameter space  \cite{cai2021exponential} because the EMA parameters are averaged from the previous iterations, but the batch-wise BN statistics are instantly collected at the current iteration. We use the recently proposed fix for this issue called EMAN \cite{cai2021exponential}, where the batch-wise statistics are exponentially averaged from the previous iterations as well. The student network is trained by solving the following optimization problem:

\begin{equation}
  \min_{\theta_S} \mathcal{L}_{student} = \mathbb{E}_{x_u\sim P_U(x)}\mathbb{E}_{\Phi\sim \mathcal{M}}[D(f(\Phi(x_u) ;\theta_{EMAN}),f(\Phi(x_u) ;\theta_S))]
\end{equation}
where $\theta_{EMAN}$ denotes the exponentially moving averaged of the teacher network's parameters from the previous iterations computed as in \cite{cai2021exponential}. $f(\Phi(x_u) ;\theta_{EMAN})$ are the pseudo-labels corresponding to $x_u$ generated using EMAN parameters.


\section{Experiments}
\subsection{Implementation Details}
The input to the networks is prepared by cropping the hand area from a depth image following  \cite{oberweger2017deepprior++} and resizing it to a fixed size of 128x128. The depth values are then normalized to [-1, 1] for the cropped image. For training, we use Adam \cite{kingma2014adam} optimizer with a cosine learning rate decay schedule \cite{loshchilov2016sgdr}. The initial learning rate is set to be $10^{-4}$, and a weight decay of $10^{-5}$ is used.  The augmentation set used for $\mathcal{M}$ includes in-plane rotation ([-180, 180] degree), 3D scaling ([0.9, 1,1]), and 3D translation ([-10, 10] mm). The proposed method uses diffident combinations of hyper parameters depending on the extent of the availability of the ground-truth labels. We refer the reader to the supplementary material for more details. We use PyTorch framework \cite{paszke2019pytorch} for implementation. 


\subsection{Datasets and Evaluation Metrics}
We evaluate the proposed method on three public 3D hand pose estimation datasets: ICVL dataset \cite{tang2014latent}, NYU dataset \cite{tompson2014real} and MSRA dataset \cite{sun2015cascaded}. The ICVL dataset contains 22K training and 1.5K testing depth images that are
captured with an Intel Realsense camera. The ground truth hand pose of each image consists of $N_j = 16$ joints. The NYU dataset is captured with three Microsoft Kinects from different views. Each view consists of 72K training and 8K testing depth images. Following most previous works, we only use the frontal view and use a subset $N_j = 14$ out of the total 36 annotated joints for training and testing in all experiments. The MSRA dataset \cite{sun2015cascaded} contains more than 76K frames captured from 9 subjects. Each subject contains 17 hand gestures and each hand gesture has about 500 frames. Each frame is provided with a ground-truth of $N_J = 21$ joints. Following the protocol used by \cite{sun2015cascaded}, we evaluate the proposed method on this dataset with the leave-one-subject-out cross-validation strategy.
\par
For evaluation, we use one of the most commonly used metrics for evaluating 3D hand pose estimation methods: the mean distance error (measured in mm). The mean distance error represents the average Euclidean distance between the estimated and the ground-truth joint locations computed over the entire testing set.

\begin{table}[t]
\caption{Performance under different strategies for adjusting the thresholds}
\centering
\setlength{\tabcolsep}{4pt}
\begin{tabular}{cc}
\hline
                    Strategy                    &   Error (mm)  \\
\hline
        {Fixed Thresholds}               & 10.43 \\
        {$\rho_t = 0.4$}                   & 9.00 \\
      {$\rho_t = 0.6$}                    & 8.84 \\
        {$\rho_t = 0.8$}                   & 8.99 \\

        {Ours}                       &   \textbf{ 8.71} \\

\hline
\end{tabular}
\label{tab:dynamic}
\end{table}

\subsection{Ablation Study}
To conduct ablation study, we choose a scenario where labeled data is scarce (more specifically, 1\% of the ground-truth labels are used) because such a scenario most highlights the proposed method's capability to leverage the unlabeled data. We use the NYU dataset for performing ablation study.
\par
\begin{figure}[t]
\centering
\includegraphics[height=3.2cm]{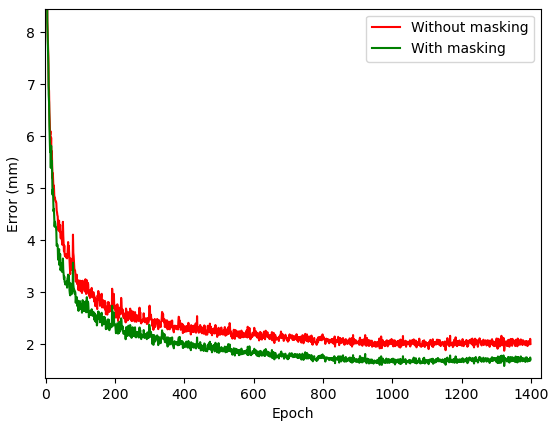}
\caption{The accuracy of pseudo-labels generated by the teacher network with and without applying masking}
\label{fig:plabels}
\end{figure}

\textbf{Impact of Sample Masking.} We study the impact of using sample masking in the training. First, we use the ground-truth labels of the unlabeled data to illustrate how the accuracy of pseudo-labels improves when we use heatmap $STD$s as the uncertainty measure to mask out noisy pseudo-labels. As shown in Fig. \ref{fig:plabels},  this approach leads to a consistent improvement of between $10\%$ to $25\%$ in the accuracy of pseudo-labels. As can be seen in Table~\ref{tab:masking}, incorporating the proposed sample masking significantly improves the performance.
\par
\textbf{Effectiveness of Dynamic Thresholding.} We examine the effectiveness of the the proposed dynamic thresholding strategy in our framework. We report the performance in three cases. In the first case, the thresholds are initialized and then kept fixed during training. The second case includes scenarios where the thresholds are dynamically adjusted according to Eq.~\ref{eqn:dynamicthreshold}, but $\rho_t$ is kept fixed. The third case refers to the proposed strategy, where $\rho_t$ is adjusted according to a cosine schedule \cite{loshchilov2016sgdr}. As can be seen in Table~\ref{tab:dynamic}, the proposed strategy leads to the best performing case.
\par


\begin{figure}[t]
\centering
\includegraphics[height=3.2cm]{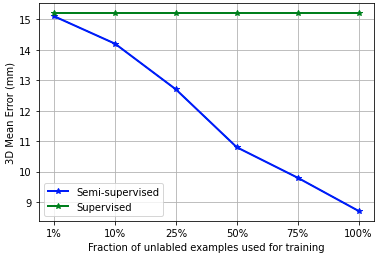}
\caption{The performance of the model under different percentages of unlabeled examples used for training}
\label{fig:semispu}
\end{figure}

\textbf{Impact of Using a Separate Network as the Student.} Empirical evidence demonstrate that using a separate network as the student (as opposed to using a single network as both the teacher and the student) leads to a performance improvement of 0.59~mm. We conjecture that this improvement primarily stems from the fact the the student network is trained using a more stationary data distribution (only the pairs of unlabeled examples and their pseudo-labels provided by the EMAN of the teacher network's parameters), as opposed to the teacher network that is trained using a combination of distributions (pairs of labeled examples and their ground-truth labels and pairs of unlabeled examples and their corresponding pseudo-labels).
\par

\textbf{Impact of Using EMAN for Generating Pseudo-Labels.} To analyse the impact of using EMAN of the teacher network's parameters (instead of the parameters themselves) for generating the training pseudo-labels for the student network, we compare the performance of the student network in three cases in terms of the parameters used for generating pseudo-labels for its training: 1) the teacher network's parameters, 2) EMA-teacher \cite{tarvainen2017mean} and 3) EMAN-teacher \cite{cai2021exponential}. As can be seen in Table~\ref{tab:params}, the best performing case is when we use EMAN. EMAN greatly improves the stability of learning by constraining the target values for the student network to change more slowly.

\begin{table}[t]
\caption{Parameters used for generating pseudo-labels for training the student}
\setlength{\tabcolsep}{4pt}
\centering
\begin{tabular}{cc}
            \hline
            Model Parameters  &   Error (mm)  \\
            \hline
                Teacher                  &         9.27 \\
                EMA-Teacher              &         10.28 \\
                EMAN-Teacher              &         \textbf{8.71} \\
        
            \hline
        \end{tabular}
        \label{tab:params}
\end{table}

\begin{table}[t]
\caption{Impact of different masking approaches on the performance}
\setlength{\tabcolsep}{4pt}
\centering
\begin{tabular}{cc}
            \hline
            Masking Approach  &   Error (mm)  \\
            \hline
                No masking          &         9.79 \\
                Binary-masking      &       9.12 \\
                Ours                &        \textbf{ 8.71} \\

        \hline
        \end{tabular}
        \label{tab:masking}
\end{table}

\subsection{Ability To Leverage Unlabeled Examples}
We analyze the effectiveness of the proposed method in leveraging unlabeled data, which is the key ability a semi-supervised method is aimed at achieving. Specifically, we use $1\%$ of the dataset as the labeled portion of the training data, and gradually expand the unlabeled portion of the training data. As can be seen in Fig. \ref{fig:semispu}, as the number of unlabeled data samples increases, the performance consistently improves. This clearly demonstrates the high capability of the proposed method of leveraging the unlabeled examples to improve its performance. 

\subsection{Comparison with State-of-the-Art Semi-Supervised Hand Pose Estimation Methods}
To demonstrate the effectiveness of the proposed method, we compare it with the state-of-the-art depth-based semi-supervised methods including \cite{chen2019so,wan2017crossing,baek2018augmented,abdi20183d}. Note that Beak et al. \cite{baek2018augmented} particularly adopt a different approach from the rest of the works. They synthesize data in the skeleton space and train a separate network to synthesize its corresponding depth image. Although they use $100\%$ of training data annotations, we include their work in our comparison as it is aimed at the same goal as the rest of the works. As can be seen in Table~\ref{table:comparision_main}, the proposed method significantly outperforms the state-of-the-art methods. Most notably, the proposed method surpasses all the existing methods when only using $1\%$ of ground-truth annotations for training. 
\par
The results from Table~\ref{table:comparision_main} also show that the proposed method enjoys a high label efficiency. Specifically, the performance of the proposed method reaches its highest level when using only $25\%$ of the ground-truth annotations. It does not achieve a considerable performance gain when more ground-truth annotations are used. Interestingly, the performance gap between the cases where $1\%$ and $100\%$ of the ground-truth annotations are used is only 0.7~mm. These observations clearly demonstrate the high capability of the proposed method of taking full advantage of the unlabeled examples. On the other hand, the existing methods rely more on the labeled data. For example, for SO-HandNet \cite{chen2019so}, the performance gap between the cases where $25\%$ and $100\%$ of the ground-truth annotations are used is 3.7~mm and 3.4~mm on NYU and ICVL respectively, indicating a very lower label efficiency.

\begin{table}[!h]
\begin{center}
\caption{Comparison of the proposed method with state-of-the-art semi-supervised methods on ICVL and NYU datasets. The performance is evaluated by the test estimation error under different percentages of labeled data used for model training
}
\label{table:comparision_main}
\setlength{\tabcolsep}{3pt}
\resizebox{\columnwidth}{!}{
\begin{tabular}{ccccc}
\hline
                    Method   &   Label Usage & Augmented Set & ICVL(mm) & NYU(mm) \\
\hline
        Beak et al.(baseline) & 100\% & No & 12.10 & 17.30 \\
        Beak et al.(w/o aug.; refine) & 100\% & No & 10.40 & 16.40 \\
        Beak et al.(w/o refine) & 100\% & Yes, 10 times & 9.10 & 14.90 \\
        Beak et al. & 100\% & Yes, 10 times & 8.50 & 14.10 \\
\hline

        \multirow{4}{*}{LSPS \cite{abdi20183d}} & 25\% & No & 7.35 & 15.70\\ 
&50\% & No & 7.10 & 15.45\\ 
&75\% & No & 7.05 & 15.45\\ 
&100\% & No & 7.00 & 15.40\\ 
\hline

        \multirow{4}{*}{Crossing Net \cite{wan2017crossing}} & 25\% & No & 10.50 & 16.10\\ 
&50\% & No & 10.0 & 16.0\\ 
&75\% & No & 10.10 & 15.90\\ 
&100\% & No & 10.20 & 15.50\\ 
\hline

        \multirow{4}{*}{SO-HandNet \cite{chen2019so}} & 25\% & No & 11.10 & 14.90\\ 
&50\% & No & 9.40 & 14.10\\ 
&75\% & No & 9.10 & 12.80\\ 
&100\% & No & 7.70 & 11.20\\ 
\hline
         \multirow{4}{*}{Ours} & 25\% & No & \textbf{6.11} & \textbf{8.14}\\ 
&50\% & No & \textbf{6.06} & \textbf{8.11}\\ 
&75\% & No & \textbf{6.04} & \textbf{8.06}\\ 
&100\% & No & \textbf{5.99} & \textbf{8.01}\\ 
\hline\hline
Ours &1\% & No & 6.94 & 8.71 \\
\hline
\end{tabular}
}
\end{center}
\end{table}
\setlength{\tabcolsep}{1.4pt}

\subsection{Semi-Supervised Learning under Extremely Low Data Regimes}
We compare the proposed method with MURAUER \cite{poier2019murauer}, which to the best of our knowledge is the only depth-based 3D hand pose estimation method examined under scenarios of severe labeled data scarcity. As can be seen in Table~\ref{tbl:comparison_low}, the proposed method significantly outperforms MURAUER \cite{poier2019murauer} under the three out of the four scenarios. The only scenario where the proposed method is inferior is when there are only 10 labeled examples. This is because such a small amount of labeled training data lacks adequate information about the nature of the task. MURAUER \cite{poier2019murauer} compensates for this lack of information by pre-training on synthetic data. However, the proposed method does not use any pre-training. More importantly, in contrast to MURAUER \cite{poier2019murauer}, the proposed method achieves this performance without the application-limiting requirement for multi-view real-data. Remarkably, in the case where there are only 100 labeled samples, the proposed method achieves 12.11~mm, which clearly demonstrates the practicality of the proposed method in cases of severely limited access to labeled data. 

\begin{table}[!h]
  \begin{center}
   \caption{Comparison of our work with \cite{poier2019murauer} under cases of severely limited access to labels on the NYU dataset. Numbers next to the methods represent their performance under the corresponding scenarios in terms of mean distance error in mm}
   \label{table:comparision_low}
   \setlength{\tabcolsep}{8pt}
  \begin{tabular}{|c|cccc|}
    \hline
    Methods & \multicolumn{4}{|c|}{Number of Labeled Examples} \\
     & 10 & 100 & 1,000 & 10,000 \\\hline

  MURAUER \cite{poier2019murauer} & \textbf{16.4} &  12.2 &  10.90 &  9.90 \\
  Ours & 25.82 & \textbf{12.11}& \textbf{8.60}&  \textbf{8.16}\\
   \hline  
  \end{tabular}
 
  \label{tbl:comparison_low}
  \end{center}
\end{table}

\begin{table}[!h]
   \caption{Comparison with the state-of-the-art fully-supervised methods on ICVL~\cite{tang2014latent} (Left), NYU~\cite{tompson2014real} (Middle), and MSRA~\cite{sun2015cascaded} (Right). “Error” indicates the mean distance error in mm. 25p and 100p respectively denote the cases where $25\%$ and $100\%$ of the ground-truth annotations are used for training}
\begin{tabularx}{\columnwidth}{CCC}
\resizebox{3cm}{!}{
    \begin{tabular}{cr}
\hline
                    Methods                    &   Error  \\
\hline
        DeepModel \cite{zhou2016model}         &        11.56 \\
      DeepPrior \cite{oberweger2015hands}      &        10.40  \\
  DeepPrior++ \cite{oberweger2017deepprior++}  &         8.10  \\
        REN-4x6x6 \cite{guo2017region}         &         7.63 \\
        REN-9x6x6 \cite{wang2018region}        &         7.31 \\
         DenseReg \cite{wan2018dense}          &         7.30  \\
         SHPR-Net \cite{chen2018shpr}          &         7.22 \\
        HandPointNet \cite{ge2018hand}         &         6.94 \\
    CrossInfoNet \cite{du2019crossinfonet}     &         6.73 \\
          NARHT \cite{huang2020hand}           &         6.47 \\
            A2J \cite{xiong2019a2j}            &         6.46 \\
       Point-to-Point \cite{ge2018point}       &         6.30  \\
        V2V-PoseNet \cite{moon2018v2v}         &         6.28 \\
          JGR-P2O \cite{fang2020jgr}           &         6.02 \\
 HandFoldingNet \cite{cheng2021handfoldingnet} &          5.95 \\
 Ours-25p                                      &        6.11 \\
 Ours-100p                                     &       5.99 \\
\hline
\end{tabular}
}
&
\resizebox{3cm}{!}{
    \begin{tabular}{cr}
\hline
                       Methods                        &   Error  \\
\hline
         DeepPrior \cite{oberweger2015hands}          &        19.73 \\
            DeepModel \cite{zhou2016model}            &        17.04 \\
                3DCNN \cite{ge20173d}                 &        14.10  \\
            REN-4x6x6 \cite{guo2017region}            &        13.39 \\
           REN-9x6x6 \cite{wang2018region}            &        12.69 \\
     DeepPrior++ \cite{oberweger2017deepprior++}      &        12.24 \\
             Pose-REN \cite{chen2020pose}             &        11.81 \\
 Generalized-Feedback \cite{oberweger2019generalized} &        10.89 \\
            HandPointNet \cite{ge2018hand}            &        10.54 \\
             DenseReg \cite{wan2018dense}             &        10.20  \\
        CrossInfoNet \cite{du2019crossinfonet}        &        10.08 \\
              NARHT \cite{huang2020hand}              &         9.80  \\
          Point-to-Point \cite{ge2018point}           &         9.10  \\
               A2J \cite{xiong2019a2j}                &         8.61 \\
    HandFoldingNet \cite{cheng2021handfoldingnet}     &         8.58 \\
            V2V-PoseNet \cite{moon2018v2v}            &         8.42 \\
              JGR-P2O \cite{fang2020jgr}              &         8.29 \\
               Ours-25p                               &        8.14 \\
              Ours-100p                                 &       8.01 \\
\hline
\end{tabular}}
&
\resizebox{3cm}{!}{
    \begin{tabular}{cr}
\hline
                    Methods                    &   Error  \\
\hline
        REN-9x6x6 \cite{wang2018region}        &         9.79 \\
             3DCNN \cite{ge20173d}             &         9.58 \\
  DeepPrior++ \cite{oberweger2017deepprior++}  &         9.5  \\
         Pose-REN \cite{chen2020pose}          &         8.65 \\
        HandPointNet \cite{ge2018hand}         &         8.5  \\
    CrossInfoNet \cite{du2019crossinfonet}     &         7.86 \\
         SHPR-Net \cite{chen2018shpr}          &         7.76 \\
       Point-to-Point \cite{ge2018point}       &         7.7  \\
        V2V-PoseNet \cite{moon2018v2v}         &         7.59 \\
          JGR-P2O \cite{fang2020jgr}           &         7.55 \\
          NARHT \cite{huang2020hand}           &         7.55 \\
 HandFoldingNet \cite{cheng2021handfoldingnet} &         7.34 \\
         DenseReg \cite{wan2018dense}          &         7.23 \\
Ours-25p                                      &          7.28 \\
 Ours-100p                                     &        7.18 \\
\hline
\end{tabular}
}

\end{tabularx}
\label{tab:comprehensive}
\end{table}
\subsection{Comparison with State-of-the-Art Fully-Supervised Methods}
We compare the proposed method with state-of-the-art methods that use $100\%$ of ground-truth annotations for training \cite{zhou2016model,oberweger2015hands,oberweger2017deepprior++,guo2017region,wang2018region,chen2020pose,oberweger2019generalized,wan2018dense,xiong2019a2j,du2019crossinfonet,fang2020jgr,ge20173d,chen2018shpr,ge2018hand,ge2018point,huang2020hand,cheng2021handfoldingnet,moon2018v2v}.
 Table~\ref{tab:comprehensive} summarizes the performance based on the mean distance error on the three datasets. As can be seen in Table~\ref{tab:comprehensive}, despite using only $25\%$ of the ground-truth annotations, the proposed method ranks as the best performing method on the NYU dataset, the second best performing on the MSRA dataset, and the third best performing on the ICVL dataset. These observations clearly demonstrate the success of the proposed framework in significantly reducing the reliance on the labeled training data. 

\section{Conclusion}
In this paper, we propose a novel framework for performing depth-based 3D hand pose estimation under scenarios where the access to the labeled data is limited. The proposed framework consists of two identical networks that are jointly trained. The teacher network is trained using consistency training on both labeled and unlabeled examples by adapting some of latest advancements in SSL methods in image classification. The student network is trained using the unlabeled examples and their corresponding pseudo-labels provided by the teacher network. Extensive experiments demonstrate the proposed framework outperforms the current state-of-the-art methods by large margins.

\bibliographystyle{unsrt}  
\bibliography{mainn}

\end{document}